\documentclass{article}
\PassOptionsToPackage{square,numbers}{natbib}
\usepackage[preprint]{neurips_2025}
\usepackage[square,numbers]{natbib}
\usepackage[utf8]{inputenc} % allow utf-8 input
\usepackage[T1]{fontenc}    % use 8-bit T1 fonts
\usepackage{hyperref}       % hyperlinks
\usepackage{url}            % simple URL typesetting
\usepackage{booktabs}       % professional-quality tables
\usepackage{amsfonts}       % blackboard math symbols
\usepackage{nicefrac}       % compact symbols for 1/2, etc.
\usepackage{microtype}      % microtypography
\usepackage{xcolor}         % colors

\usepackage{graphicx}
\usepackage{amsmath}
\usepackage{subcaption}
\usepackage{caption}
\usepackage{threeparttable}
\usepackage{multirow}
\usepackage{pifont}
\usepackage{wrapfig}
\usepackage{algorithm}
\usepackage{algorithmic}
\usepackage[toc,page]{appendix}
\usepackage{tabularx}
\usepackage{cprotect}
\usepackage[dvipsnames]{xcolor}
\usepackage{listings}
\usepackage[most]{tcolorbox}
\tcbuselibrary{listings}
\usepackage{wrapfig}

\NewTCBListing{promptbox}{ !O{} }{listing only, #1}

\title{Synthelite: Chemist-aligned and feasibility-aware synthesis planning with LLMs }

\author{Xuan Vu Nguyen \textsuperscript{1}, \quad
  Daniel Armstrong\textsuperscript{1}, \quad Milena Wehrbach\textsuperscript{1}, \\
  \textbf{\quad Andres M Bran\textsuperscript{1,2,3}, \quad Zlatko Jončev\textsuperscript{1,3}, \quad Philippe Schwaller\textsuperscript{1,2}} \\
  \textsuperscript{1}\'Ecole Polytechnique F\'{e}d\'{e}rale de Lausanne (EPFL) \\
  \textsuperscript{2}National Centre of Competence in Research (NCCR) Catalysis \\
  \textsuperscript{3}B12 Labs \\
  \texttt{\{nguyen.nguyen,philippe.schwaller\}@epfl.ch} \\
}
\begin{document}

\maketitle

\begin{abstract}
  Computer-aided synthesis planning (CASP) has long been envisioned as a complementary tool for synthetic chemists.
  However, existing frameworks often lack mechanisms to allow interaction with human experts, limiting their ability to integrate chemists' insights.
  In this work, we introduce Synthelite, a synthesis planning framework that uses large language models (LLMs) to directly propose retrosynthetic transformations.
  Synthelite can generate end-to-end synthesis routes by harnessing the intrinsic chemical knowledge and reasoning capabilities of LLMs, while allowing expert intervention through natural language prompts.
  Our experiments demonstrate that Synthelite can flexibly adapt its planning trajectory to diverse user-specified constraints, achieving up to 95\% success rates in both strategy-constrained and starting-material-constrained synthesis tasks.
  Additionally, Synthelite exhibits the ability to account for chemical feasibility during route design.
  We envision Synthelite to be both a useful tool and a step toward a paradigm where LLMs are the central orchestrators of synthesis planning.
\end{abstract}

\section{Introduction}
% This is something I just added as a general intro, I don't mind if you cut it out or have a different picture of the red line
In organic chemistry, \textit{retrosynthesis} refers to the stepwise deconstruction of a complex molecule into accessible precursors and stands among the most prestigious and intellectually refined disciplines of the field \cite{corey1969computer,corey1991logic}. 
Like many decision-making problems, the search space grows exponentially with the number of synthetic steps.
Automated synthesis planning has therefore relied heavily on explorative search algorithms \cite{chen2020retro,segler2018planning,schwaller2020predicting,browne2012survey},  combined with data-driven or heuristic single-step policies \cite{modelsmatter,schwaller2019molecular,tu2022permutation}.
Although these approaches excel at generating synthetic routes that map simple target molecules to available starting materials, they remain limited as practical assistants for synthetic chemists.
In practice, expert chemists frequently impose preferences on their synthesis design, such as the order of reactions \cite{shenvi2009chemoselectivity} (e.g. performing high-risk reactions early) or the choice of starting materials \cite{lopez2024application,wolos2022computer} (e.g. prioritizing waste valorisation).
However, given the vast solution space, finding a synthesis route that aligns with such preferences is rarely guaranteed, often requiring iterative exploration through numerous candidates.
Recent work on constrained synthesis planning has begun to incorporate user-specified constraints, such as bond cleavage preferences \cite{westerlund2025human,thakkar2023unbiasing} or predefined starting materials \cite{armstrong2025tango,yu2024double}.
Yet, a more general framework that can flexibly accommodate arbitrary constraints through natural language prompting remains absent.
% However, a more universal framework that allows arbitrary constraints via textual prompting is still lacking.
% It combines art and science, demanding a deep understanding of chemical reactivity, strategic reasoning, and synthetic intuition.\cite{corey1969computer,corey1991logic}\\ \\
% While computer-aided synthesis planning has made significant progress in the past years \cite{}, most approaches to date still opt for synthetic feasibility rather than synthetic strategy, missing out on the chemical intuition when designing models. Thus, the nuanced choices that define expert-level planning, like in which order to form rings or when to introduce labile functional groups, often lie beyond the scope of current algorithms. \\ \\

Recent advancements in large language models (LLMs) \cite{ouyang2022training,vaswani2017attention} and chain-of-thought reasoning \cite{guo2025deepseek,snell2024scaling,muennighoff2025s1} have enabled these models to tackle increasingly complex logical tasks. They can now produce coherent multi-step solutions for problems that traditionally rely on exploratory search algorithms, such as Sudoku or chess \cite{giadikiaroglou2024puzzle,feng2023chessgpt}.
Moreover, the expansion of LLMs training into scientific domains has opened new opportunities for their application in research and discovery \cite{zimmermann2024reflections,bran_chemcrow_2023,zheng2025automation}.
In particular, recent studies demonstrate that LLMs possess a growing understanding of small-molecule chemistry, exhibiting fine-grained reasoning over functional groups and reaction patterns \cite{bran2025chemical,mirza2024large,chen2023chemist,qian2023can}.

\begin{figure}[t]
    \centering
    \includegraphics[width=1\textwidth]{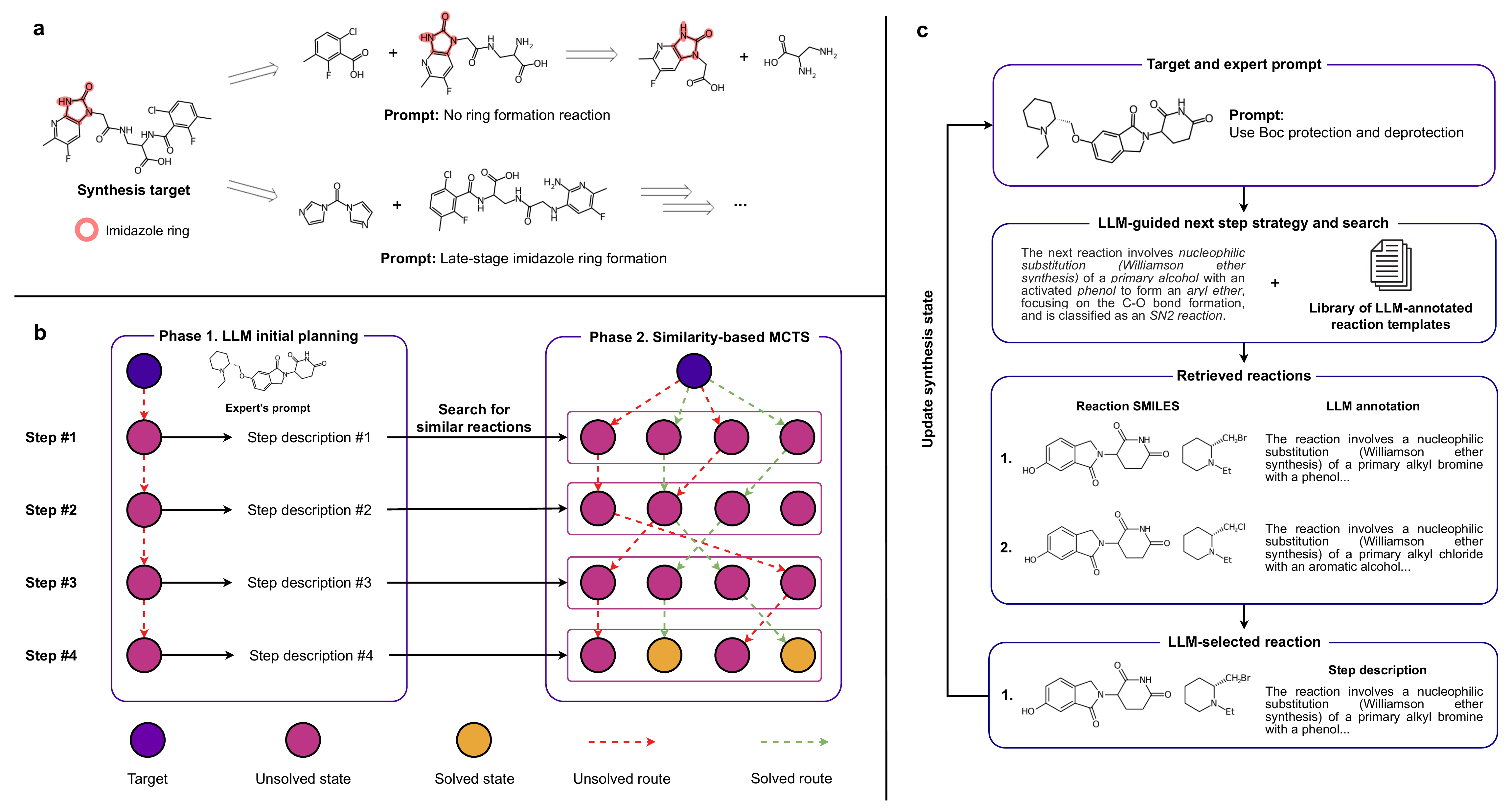}
    \caption{\textbf{a) }An example of the task of expert-prompted synthesis planning. \textbf{b) } Conceptual overview of Synthelite, consisting of two phases: In Phase 1, the LLM drafts a strategy for each reaction step; in Phase 2, searching for a combination of reactions that can lead the synthesis to in-stock materials while aligned with the LLM-proposed strategy in Phase 1. \textbf{c) } A close-up on Phase 1. For each step of the synthesis, the LLM produces a textual description of the next retro step, which is then used to retrieve relevant reactions from an LLM-annotated template database. The LLM then chooses the most suitable reaction to update the synthesis.}
    \label{fig:concept}
\end{figure}

Initial works have been done to incorporate LLMs into CASP frameworks, using LLMs either as reaction validator \cite{bran2025chemical,baker2025larc} or few-shot multi-step predictor \cite{wang2025llm,song2025aot}.
In this work, we introduce Synthelite, an LLM-driven synthesis planning framework in which large language models serve as the central agents for route generation and decision-making. 
Our contribution is as follows:
\begin{enumerate}
    \item We propose a framework that lets the LLM perform retrosynthetic planning step-by-step, leveraging its intrinsic chemical knowledge and reasoning capabilities. 
    By using the LLM as the backbone, Synthelite naturally accommodates user input through textual prompts, allowing flexible incorporation of expert constraints. Our results show that Synthelite can generate synthesis plans that not only align closely with expert-provided guidance but also exhibit distinct strategic variations for the same target under different prompts.
    % \item We shows that Synthelite can handle diverse constrains from human chemists, ranging from strategy to starting material constrains. 
    \item We show that the latent chemical knowledge embedded in modern LLMs allows Synthelite to account for practical, wet-lab feasibility when designing synthetic routes, bridging the gap between computational tools and experimentation.
    \item We demonstrate that our framework enables LLMs to achieve an improved solve rate on retrosynthesis benchmarks within a reasonable request budget.
\end{enumerate}
% We propose a framework that lets the LLM perform retrosynthetic planning step-by-step, leveraging its intrinsic chemical knowledge and reasoning capabilities. 
% % Then, in another phase, Monte Carlo tree search (MCTS) algorithm using a similarity-base policy is employed to map the LLM-proposed routes to routes that lead to in-stock molecules.
% % In a subsequent phase, a Monte Carlo Tree Search (MCTS) algorithm with a similarity-based policy refines these LLM-proposed routes, mapping them to viable pathways that terminate in commercially available starting materials (Figure \ref{fig:concept}).
% By using the LLM as the backbone, Synthelite naturally accommodates user input through textual prompts, allowing flexible incorporation of expert constraints. Our results show that Synthelite can generate synthesis plans that not only align closely with expert-provided guidance but also exhibit distinct strategic variations for the same target under different prompts.
% Moreover, we show that the chemical knowledge embedded in modern LLMs allows Synthelite to account for practical, wet-lab feasibility when designing synthetic routes.

\section{Related work}
\subsection{Neural-guided synthesis planning}

The predominant paradigm in CASP combines a neural policy model that maps a target molecule to its immediate precursors \cite{localretro,schwaller2020predicting,irwin2022chemformer,tu2021permutation}, which is then iteratively invoked within an explorative search algorithm, such as Monte Carlo Tree Search (MCTS) \cite{segler2018planning,swiechowski2023monte} or A*-inspired methods \cite{chen2020retro}.
The policy models are typically trained on single-step reactions, assuming that multi-step retrosynthesis is a Markov process. However, this assumption departs significantly from practical organic synthesis, where reaction steps often exhibit long-range dependencies. For instance, a deprotection step may be required to remove a protecting group that was introduced several steps earlier.

Beyond such dependencies, practical synthesis planning also incorporates higher-level strategic considerations aimed at improving the likelihood of wet-lab success. 
These factors include convergent synthesis that shortens the longest path in the synthesis tree, stability of intermediates \cite{corey1991logic,beller2022vicinal}, and sustainability \cite{weber2021chemical,sheldon2017factor,anastas2000green}.
Nevertheless, most existing approaches primarily optimize the search based on solve rate - namely, the ability to reach purchasable starting materials at the leaf nodes - while largely overlooking strategic reasoning and overall route quality. Furthermore, these systems offer limited user steerability. Although recent methods introduce a certain level of controllability, such as disconnection constraints via bond tagging \cite{westerlund2025human} or constraints on building blocks \cite{armstrong2025tango,yu2024double}, a fully interactive, omni-conditional synthesis planning framework remains absent.

\subsection{LLMs as single-step policies}
% Cite Alan's work
Several works have explored training reasoning-capable LLMs to serve as single-step policies in place of the traditional neural-based ones \cite{narayanan2025training,zhang2025reasoning}, with the notion that chain-of-thought reasoning would offer strategic consideration and explainability to the proposed reactions. However, an inherent limitation of LLMs is their consistency in producing valid symbolic representations, such as SMILES or SMARTS strings \cite{edwards-etal-2022-translation,walters2024silly,jang2024}. To mitigate this issue, \citet{hassen2025atom} proposes a structured reasoning paradigm that first identifies reaction sites and leverages a reaction ontology to ground LLM outputs into valid chemical transformations. Nonetheless, a more fundamental limitation of LLM-based policies is their high computational cost, which renders them impractical for direct integration into explorative search algorithms.

\subsection{LLMs as validators for synthesis steering}

Given the high computational cost of LLMs as search policies, several works instead employ them as scoring or evaluation modules, either at the route level \cite{bran2025chemical} or the reaction level \cite{baker2025larc}, on top of conventional synthesis planning systems. While this configuration enables natural-language input from users, the overall synthesis trajectory remains largely governed by the underlying algorithm, thereby limiting the degree of steering and restricting exploration of novel solution spaces. Moreover, depending on the imposed constraints, the search space could be highly sparse, in which case the existence of a satisfactory synthesis route is not guaranteed.

\subsection{Multi-step planning with few-shot prompting}
Another line of work augments LLMs with example synthesis routes of molecules that are structurally similar to the given target \cite{song2025aot,wang2025llm}. In these approaches, the LLM adapts retrieved routes to the target molecule, in conjunction with graph-based exploration or evolutionary optimization to reach purchasable starting materials. However, such methods rely heavily on the quality and relevance of the retrieved examples and may suffer from synthesizability cliffs, wherein reaction pathways suitable for one molecule may fail to transfer to its analogues. Moreover, similar to scorer-based approaches, dependence on known synthetic procedures limits the ability of these frameworks to explore new solutions, and example routes that align with user-imposed constraints are not always available.
%%%

In this work, we aim to explore the direction of using LLMs as the central agent in synthesis planning, without bias from few-shot samples and expensive policy calls.

\section{Method}
Synthelite is a two-phase synthesis planning framework that integrates LLMs with a similarity-based MCTS. 
To mitigate LLMs’ limitations in generating valid symbolic chemical representations \cite{edwards-etal-2022-translation, walters2024silly, jang2024}, we first construct a template search engine based on AiZynthFinder’s library of ~40,000 SMARTS-encoded reactions \cite{genheden2020aizynthfinder}. Each template is converted into a one-sentence textual description by an LLM and embedded into a vector space to enable semantic similarity search. 
In Phase 1, the LLM (\verb|Claude-4.5-Sonnet|, \verb|Gemini-2.5-Pro|, or \verb|GPT-5|) iteratively plans a synthesis route from the target molecule toward available starting materials, describing each retro-step in text. These descriptions are then matched to reaction templates, and the LLM selects the best-fitting reaction, forming a blueprint route. In Phase 2, MCTS refines this blueprint by locally sampling reactions similar to those proposed in Phase 1, guided by a weighted scoring function balancing alignment with LLM-proposed strategies and template popularity, and optimized toward the stock availability of the leaf molecules.
To enable some diversity and self improvement, the Phase 1 is run three times, with subsequent attempts informed with self-generated feedback from the previous ones.
For more details about the methodology, see Appendix \ref{sec:method}.

\section{Results \& Discussion}
\subsection{Strategic synthesis planning}

\begin{figure}
    \centering
    \includegraphics[width=0.93\linewidth]{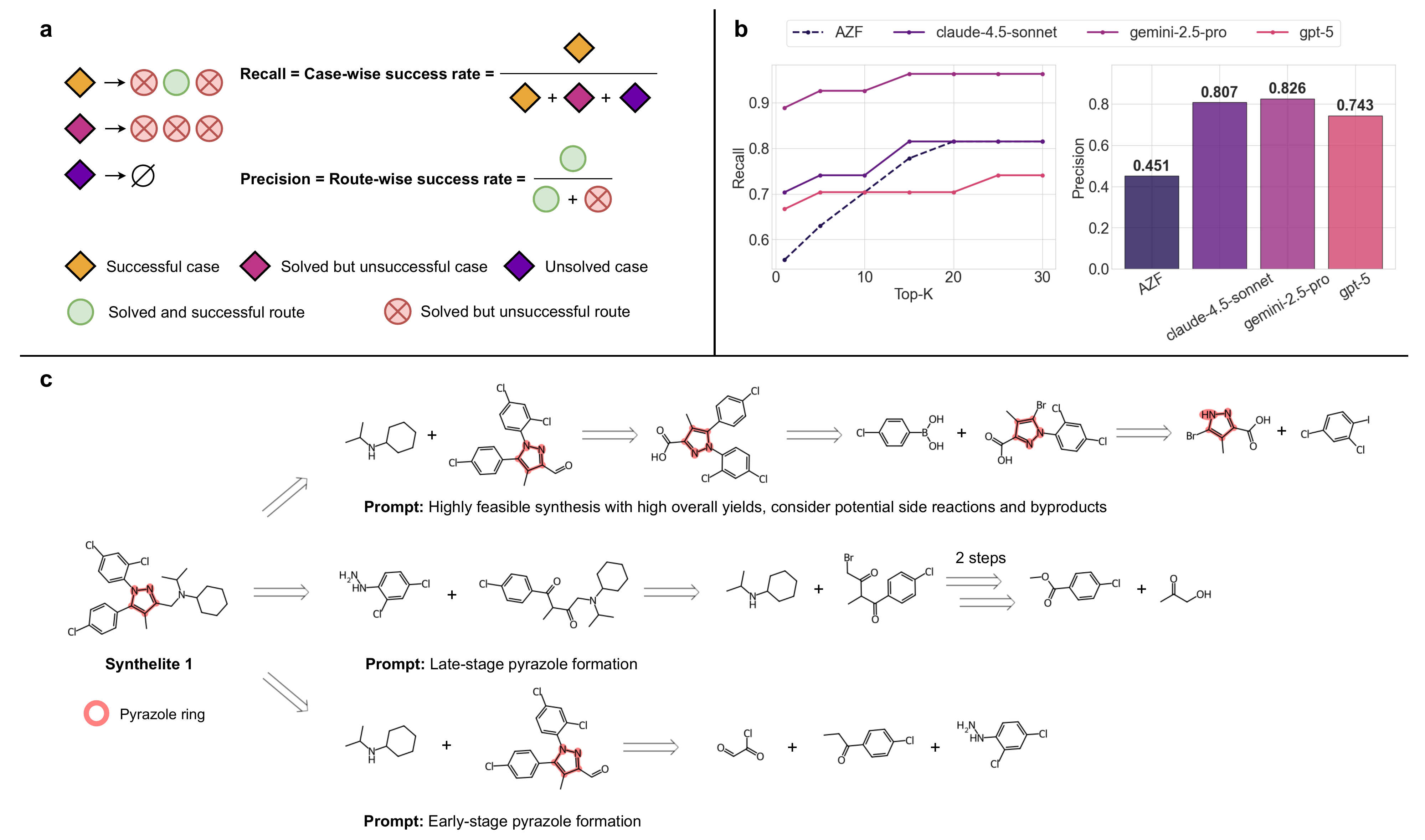}
    \cprotect\caption{\textbf{a) }Metric definition: A pair of target and prompt is considered to be solved successfully if the model can find at least 1 route that passes the validation scripts. \textbf{b) }Precision and recall of Synthelite, varying the LLM, compared with the AZF baseline.
    \textbf{c) }An example showing how Synthelite can adapt to different expert prompts. Routes are produced using \verb|Claude-4.5-sonnet|.}
    \label{fig:main_result}
\end{figure}

To assess Synthelite’s ability to follow user-provided prompts, we extend upon the benchmark introduced by \citet{bran2025chemical}, which includes pairs of potent drug molecules developed by pharma and universities, along with corresponding steering prompts (see Section \ref{sec:benchmark}). Our extended benchmark comprises eight targets spanning different levels of synthetic complexity, each associated with several textual prompts, resulting in a total of 27 molecule–prompt pairs. 
Each test case is associated with a script to check if the user query is satisfied.
As a baseline, we run AiZynthFinder (AZF) \cite{genheden2020aizynthfinder} on all eight targets without additional inputs.

% To evaluate how well Synthelite is at following user prompts, we extend upon the benchmark by \citet{bran2025chemical}, containing pairs of targets and steering prompts (see Section \ref{sec:benchmark}).
% Our final benchmark contains 8 targets with various levels of synthetic complexity, each is associated with a number of prompts written in textual format, for a total of 27 pairs of molecule-prompt. As a baseline, we run AiZynthFinder (AZF) \cite{genheden2020aizynthfinder} on all eight targets without additional inputs.

The alignment between the generated routes and expert prompts is evaluated using recall and precision metrics (Figure~\ref{fig:main_result}a, b). Recall quantifies the proportion of successful cases in which at least one valid route satisfying the prompt is found among the top-K samples. Owing to its exploratory nature, AiZynthFinder (AZF) can occasionally identify solutions that meet expert requirements by chance. This effect is reflected in the increase of recall with larger sample sizes: at top-1, AZF successfully solves 55\% (15/27) of the test cases, and it requires 20 samples to reach 81\% (22/27).
In contrast, Synthelite exhibits a more concentrated search profile, as indicated by its flatter recall curves, often identifying a correct route at top-1. Among the tested LLMs, \verb|Gemini-2.5-Pro| achieves the highest recall, with 89\% (24/27) case-wise success at top-1 and 95\% (26/27) at top-30,  outperforming AZF across all sample sizes. \verb|Claude-4.5-Sonnet| and \verb|GPT-5| follow similar trends, with the former maintaining higher recall than AZF up to top-20. Although \verb|GPT-5| is the least performant among the LLMs tested, it still surpasses AZF with fewer than ten sampled routes.
We note that this performance profile is still advantageous in practical settings, where users are typically interested only in the top few routes suggested by a synthesis planning system.

To assess the quality of the suggestions, we measure precision, defined as the proportion of generated routes that satisfy the user prompt. As shown in Figure~\ref{fig:main_result}b, all LLM-based variants of Synthelite exhibit higher precision than AiZynthFinder (AZF), indicating that Synthelite’s search space is more focused on solutions aligned with expert guidance. Among the tested models, Synthelite powered by \verb|Gemini-2.5-Pro| achieves the highest precision at 83\%, closely followed by \verb|Claude-4.5-Sonnet| and \verb|GPT-5|.

% To characterize the suggestion quality, we opt to measure precision, i.e. how frequently the model suggests routes that satisfy the prompt. Figure \ref{fig:main_result}b shows that all LLM-variants of Synthelite has higher precision compared with AZF, highlighting that the search space of Synthelite is more focused on solutions that aligned with the expert prompts.
% Synthelite with \verb|Gemini-2.5-Pro| has the precision as high as 83\%, closely followed by \verb|Claude-4.5-Sonnet| and \verb|GPT-5|.

Figure~\ref{fig:main_result}c illustrates how Synthelite adapts its retrosynthetic strategy to synthesize the target molecule \textbf{Synthelite 1}, which features a central pyrazole ring, in response to different prompts. Under a strategy-neutral instruction (Figure \ref{prompt:neutral}), \verb|Claude-4.5-Sonnet| generates a plan that avoids explicit pyrazole ring formation, instead prioritizing well-established transformations such as reductive amination or Suzuki coupling \cite{miyaura1995palladium}. When prompted to include a pyrazole formation reaction, however, Synthelite adjusts its plan accordingly. For example, when asked for “Late-stage pyrazole formation”, it proposes a pathway in which the final step involves a Knorr pyrazole synthesis reaction \cite{knorr1883einwirkung}. Conversely, when instructed to perform “Early-stage pyrazole formation”, Synthelite restructures its route to begin with a reductive amination step followed by a multicomponent ring condensation reaction \cite{betcke2024multicomponent}.

\subsection{Starting material constrained synthesis}
\begin{wrapfigure}{r}{0.45\textwidth}
  \centering
  \includegraphics[width=\linewidth]{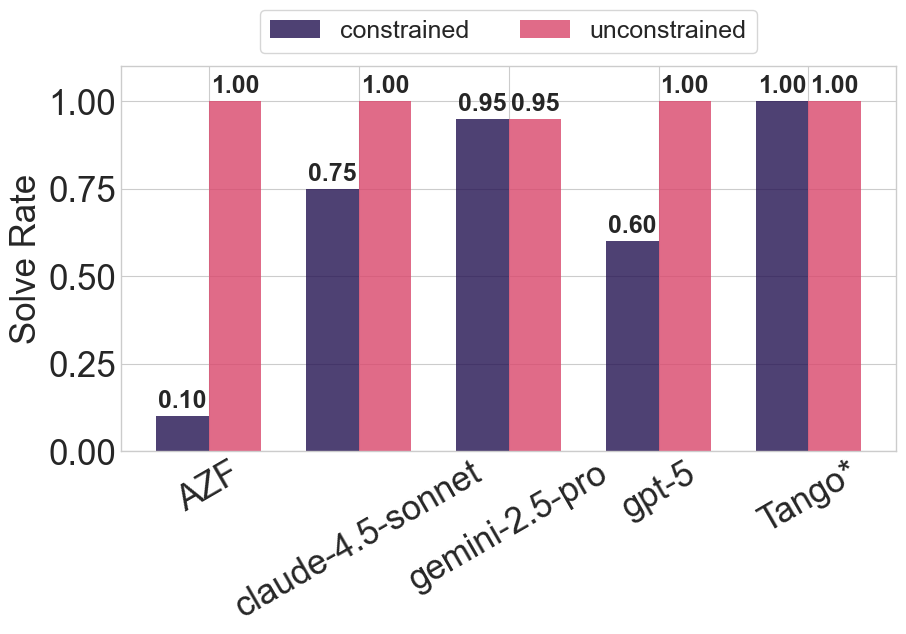}
  \caption{Solve rate on a subset of 20 targets from Pistachio Reachable, with and without starting material constraint.}
  \label{fig:sm_constrained}
\end{wrapfigure}

In starting materials constrained synthesis, besides the target molecule, the input also includes a leaf molecule that the synthesis pathway must end up with \cite{armstrong2025tango,yu2024double}.
The task is relevant in the context of semi-synthesis, where a synthetically challenging motif in the target can be readily derived from commercially available building blocks \cite{ojima1992new}, or valorisation of industrial waste \cite{wolos2022computer,zkadlo2024computational}.
We apply Synthelite to this task by simply providing the IUPAC names and SMILES strings of the required leaf molecules in the prompt and instructing the LLM to incorporate them as starting materials (see Figure \ref{prompt:sm_constraint}).
For benchmarking, we randomly select 20 targets from the 140 targets in Pistachio Reachable \cite{yu2024double}.

We consider this task to be more challenging than the strategy-constrained setting, as success can only be verified at the completion of the full synthesis route.
The sparsity of the success landscape is evidenced by the performance gap between constrained and unconstrained settings for AiZynthFinder (Figure \ref{fig:sm_constrained}). While AiZynthFinder successfully solves all 20 targets in the unconstrained setting, it identifies a solution for only one target when a specific starting material is imposed.
In contrast, Synthelite paired with frontier LLMs exhibits stronger alignment with the constrained objective: \verb|Gemini-2.5-Pro| achieves the highest solve rate among the evaluated LLMs at 95\% (19/20), followed by \verb|Claude-4.5-Sonnet| and \verb|GPT-5| with solve rates of 75\% (15/20) and 60\% (12/20), respectively. For reference, Tango* \citep{armstrong2025tango}, a search-based approach equipped with a dedicated value function to steer the search toward the desired starting material, attains a 100\% constrained solve rate.
These results indicate that even in the absence of an explicit value function tailored to starting-material constraints, the reasoning capabilities of LLMs, particularly \verb|Gemini-2.5-Pro|, can effectively guide synthesis planning toward specified building blocks.

\subsection{Feasibility-aware synthesis planning}

\begin{figure}[t]
    \centering
    \includegraphics[width=0.99\linewidth]{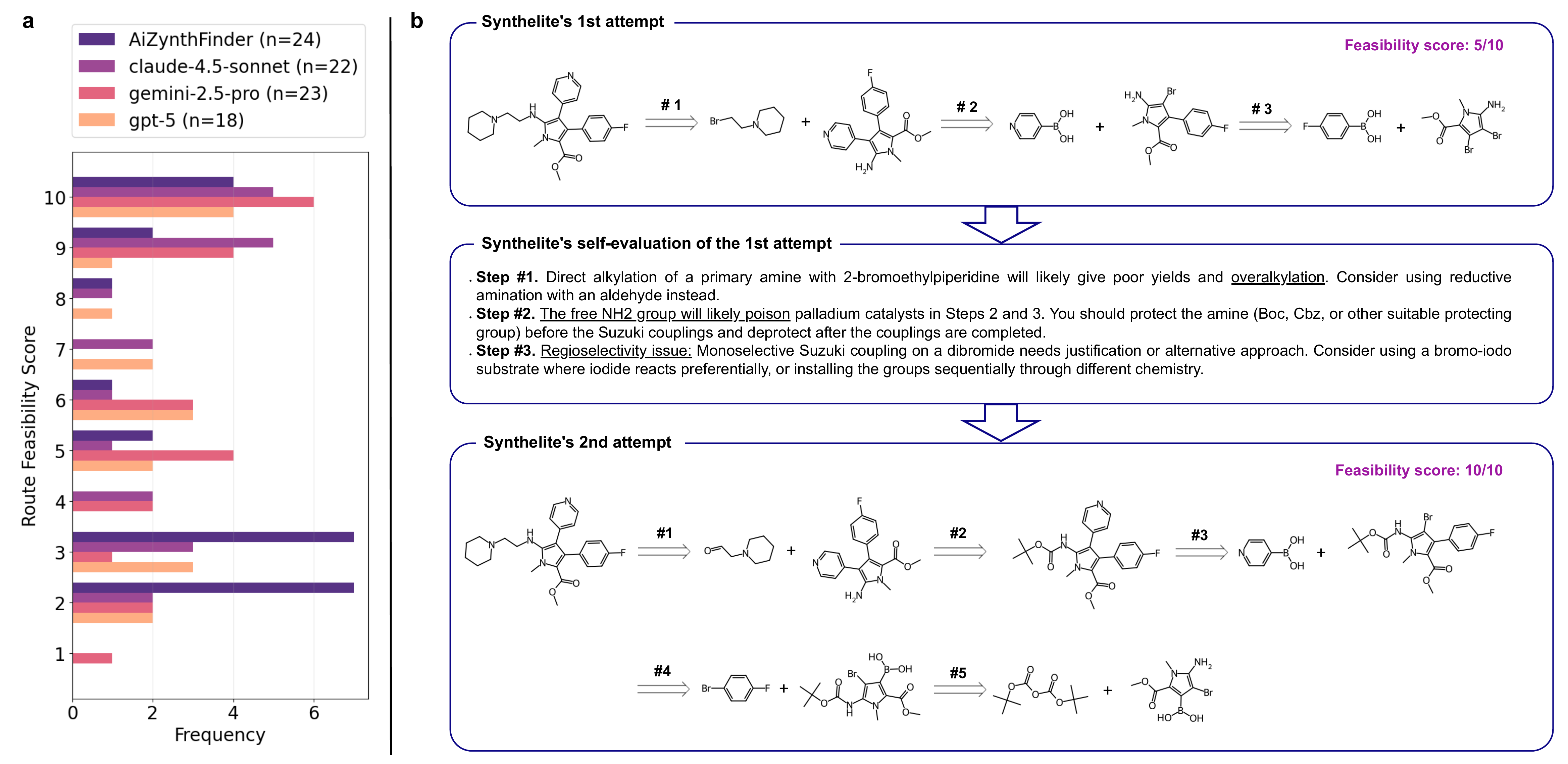}
    \cprotect\caption{a) Distribution of feasibility scores for the top-3 routes across eight targets, evaluated by \verb|Gemini-2.5-Pro|. b) Example illustrating the chemical feasibility awareness of the Synthelite framework, where the LLM refines its synthesis strategy across successive attempts. Example shown for target Synthelite 2, with routes generated by \verb|Claude-4.5-Sonnet|.}
    \label{fig:feasibility}
\end{figure}

As highlighted by \citet{bran2025chemical}, recent frontier LLMs have demonstrated strong capabilities in synthetic chemistry analysis, exhibiting fine-grained reasoning about selectivity, reaction conditions, expected yield, and other aspects of chemical reactivity, given only the SMILES representations of reactants and products. 
To investigate how such knowledge might enhance the quality of synthesis planning, we collected the top three routes proposed by Synthelite for each target and employed \verb|Gemini-2.5-Pro| as an evaluator. The judge model was prompted following the same protocol as \citet{bran2025chemical}, assigning a score from 1 to 10 to each route, where 1 denotes a low-yield or implausible pathway and 10 corresponds to a highly feasible route likely to succeed under experimental conditions. 
The motivation behind using the highest score instead of the average one is that we observe the LLM-judge would give a route low score if there is \textit{at least} one reaction in the pathway is bad, which makes sense since the yield of a route is only as good as that of the weakest link.
However, \verb|Gemini-2.5-Pro| does not always correctly interprete a reaction SMILES, leading to false negative evaluation.
Therefore, in a optimistic setting, we evaluate the feasibility of each route three times and take the highest score among the three runs as the final score.
To ensure objectivity, Synthelite was prompted with a strategy-neutral instruction for all targets, avoiding prompt-induced bias (Figure \ref{prompt:neutral}).

The distribution of feasibility scores is shown in Figure \ref{fig:feasibility}a. Across all LLMs, routes generated by Synthelite are consistently skewed toward higher scores compared with those produced by AZF, many of which receive only 2 or 3 out of 10.
Figure \ref{fig:feasibility}b illustrates how Synthelite incorporates chemical feasibility into its synthesis design process.
For each target, the LLM is given three planning attempts, with each subsequent attempt informed by self-generated feedback from the previous one. This feedback includes evaluations of individual reactions and suggestions for improvement.
In the example of target \textbf{Synthelite 2}, which features three interconnected aromatic rings, \verb|Claude-4.5-Sonnet| initially proposes an amination of a bromide-substituted substrate, followed by two consecutive Suzuki couplings.
Upon self-evaluation, the model identifies three key issues in the first attempt: (1) the amination of a halide could result in over-alkylation, forming a tertiary instead of a secondary amine, (2) the unprotected primary amine might poison the Pd catalyst \cite{zhang2024suzuki}, (3) and the Suzuki coupling at step \textbf{\#3} involves a substrate with two bromides, which could lead to poor selectivity.
These analysises align with the external feasibility assessment by the judge, which assigns this route a moderate score of 5/10.
In the second attempt, the LLM replaces the halide amination with reductive amination of an aldehyde to prevent over-alkylation, introduces Boc protection of the amine to avoid catalyst poisoning, and interchanges the bromide and boronic acid in step \textbf{\#4} to improve selectivity.
This revised route is judged experimentally plausible by \verb|Gemini-2.5-Pro|, which assigns a high feasibility score of 10/10. 
We envision that in the practical settings, the inter-attempt feedback could also be provided by users.

% All LLM variants solve all 8 targets, except for \verb|GPT-5| could not solve \textbf{Steer 4}

% Potentially good case studies:
% * Steer 3 of Gemini, that learns to change the direction based on feedback from previous attempt
% * Synthelite 2 of Claude. It proposes protection reaction of amino group so that it doesn't poison the Pd catalyst of Suzuki coupling.

% \subsection{Case study: Synthelite can propose strategies differing from AZF}

% Mentioning the route generation of AZF is governed by single-step reaction distribution without regarding the context of the synthesis, while Synthelite constructs routes by planning multi-steps ahead.

% Show Steer 3 and its feasibility scores distribution maybe? Analyze focusing on multi-step planning.

% Put the others to appendix.

\subsection{Ablation study}
\begin{figure}
    \centering
  \includegraphics[width=0.6\linewidth]{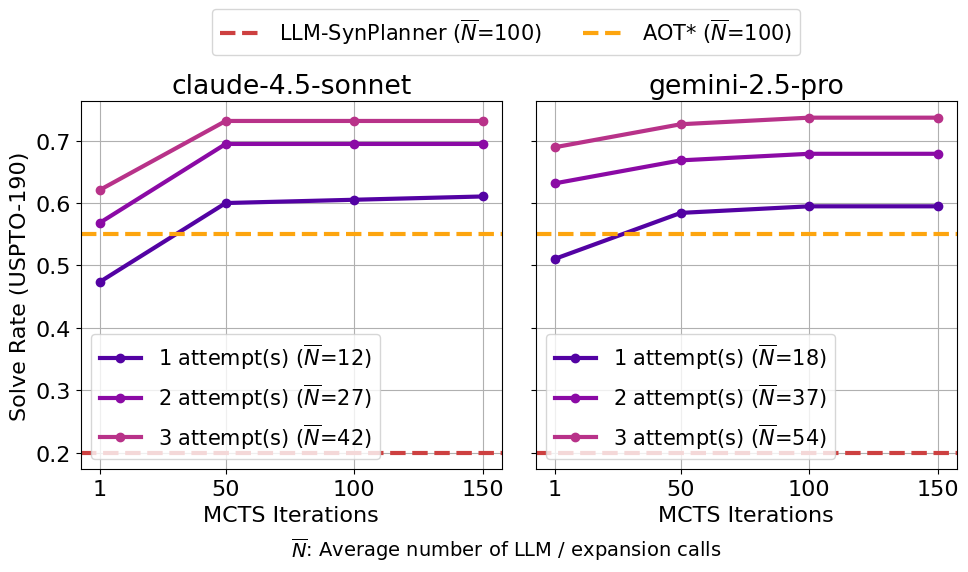}
  \caption{The effect of multi-attempt Phase 1 and the Phase 2's MCTS on the solve rate of Synthelite on USPTO-190.}
  \label{fig:uspto-190}
\end{figure}

% \begin{wrapfigure}{r}{0.5\textwidth}
%   \centering
%   \includegraphics[width=\linewidth]{figs/uspto_190.png}
%   \caption{The effect of multi-attempt Phase 1 and the Phase 2's MCTS on the solve rate of Synthelite on USPTO-190.}
%   \label{fig:uspto-190}
% \end{wrapfigure}

To justify our design choice for Synthelite, Figure \ref{fig:uspto-190} shows the effect of multiple planning attempts in Phase 1 and Phase 2's MCTS.
Results are shown for the solve rates on the USPTO-190 benchmark \cite{chen2020retro}, which comprises 190 challenging retrosynthesis targets curated from the USPTO database \cite{Lowe2012}.
When LLMs are only allowed to generate synthesis plans greedily, without retries or MCTS, \verb|Claude-4.5-Sonnet| and \verb|Gemini-2.5-Pro| achieve solve rates of 47\% and 51\%, respectively. Allowing each model to retry two additional times, while still disabling Phase 2, already improves the solve rate to 62\% for \verb|Claude-4.5-Sonnet| and 69\% for \verb|Gemini-2.5-Pro|.
% Simply letting the models retry two more times (without Phase 2), the solve rates are already boosted up to 62\% for \verb|Claude-4.5-Sonnet| and 69\% for \verb|Gemini-2.5-Pro|.
For each planning attempt, Phase 2 applies MCTS to refine the greedy routes produced in Phase 1 into the route space that maps to commercially available starting materials. With a budget of 150 MCTS iterations, we consistently observe further gains in solve rate across all attempts. The search quickly reaches a plateau as the exploration is constrained to the strategies of the LLMs.
Combining multiple planning attempts with MCTS refinement enables Synthelite to reach an overall solve rate of approximately 74\% with both \verb|Gemini-2.5-Pro| and \verb|Claude-4.5-Sonnet|.
For reference, AOT* \cite{song2025aot} and LLM-SynPlanner \cite{wang2025llm} report zero-shot solve rates of 55\% and 20\%, respectively. We note, however, that this comparison is just an estimation, as these baselines rely on a different LLM, namely DeepSeek \cite{guo2025deepseek}.

%%%

\section{Conclusion}
In this work, we introduce Synthelite, an explorative inference framework for large language models that integrates their latent strategic reasoning and chemical knowledge with Monte Carlo Tree Search to address the retrosynthesis problem. This is achieved by decoupling LLM inference and MCTS into two distinct phases, therefore avoiding expensive policy-level LLM calls during search. Moreover, LLM hallucinations are mitigated by allowing models to give reactions in natural language, bypassing rigid symbolic outputs.

Our analysis indicates that current frontier LLMs are capable of planning multi-step retrosynthetic routes while accounting for experimental plausibility and optional expert guidance, all while offering explainability through explicit reasoning traces.
The use of natural language prompting further enables seamless human interaction and the incorporation of omni-conditional inputs. Across both the strategy-constrained and starting-material-constrained benchmarks, Synthelite demonstrates strong controllability, with \verb|Gemini-2.5-Pro| in particular achieving high success rates in both settings.
We further show that, when wet-lab feasibility is treated as an implicit constraint, LLMs can reason about factors such as chemoselectivity, catalyst poisoning, and protection–deprotection strategies to improve overall route quality.
Finally, our ablation study demonstrates that, within a reasonable request budget, letting LLMs to iteratively replan with awareness of prior attempts, combined with subsequent MCTS, can effectively improve the solve rates of Synthelite on the USPTO-190 benchmark in comparison with greedy inference.

%%%

While LLMs offer a number of benefits, the current reliance on closed-source models introduces practical challenges, particularly in reproducibility, accurately measuring runtime, and conducting fair comparisons with existing retrosynthesis methods. In addition, we observe a recurrent failure mode stemming from limitations of the reaction template library: in some cases, reactions proposed by the LLM cannot be matched to available templates, leading to premature termination of the planning process. Consequently, future work will focus on reducing dependence on closed-source LLMs and on developing more efficient mechanisms for LLM-guided single-step retrosynthesis. Despite these limitations, we believe that Synthelite still represents a meaningful step toward bridging LLM-based reasoning with the inherently explorative nature of synthesis planning, moving closer to a more practical and transparent computer-aided synthesis planning tool.

% Outlooks
% * Less reliant on close-source LLMs.
% * Less reliant on a limited template library.
% * reduce the number of requests for each step - 

% Frontier LLMs are now able to plan a synthesis from end-to-end.
% Synthelite offers a framework that helps the LLMs to overcome their limitation in symbolic writing

% Better quality control over the template library.

% Increase benchmark size and prompt diversity, expanding to starting material constraint and reaction constraint

% Benchmark solve-rate on common benchmarks like USPTO-190 \cite{chen2020retro}, or SimpRetro \cite{li2024challenging}

\section*{Acknowledgement}
X.V.N acknowledges the support from the AiChemist project via MSCA Doctoral Network.
D.A and Z.J acknowledge the support by the Swiss National Science Foundation (SNSF) (grant number 214915).
A.M.B and P.S acknowledge the support from the NCCR Catalysis (grant number 225147).

\section*{Data and code availability}
Source code and benchmarks can be found at \href{https://github.com/schwallergroup/synthelite}{https://github.com/schwallergroup/synthelite}.
% Synthesis pathways produced by Synthelite can be found at ...

\bibliography{ref}

%%%%%%%%%%%%%%%%%%%%%%%%%%%%%%%%%%%%%%%%%%%%%%%%%%%%%%%%%%%%

\clearpage
\appendix
\counterwithin{figure}{section}
\counterwithin{table}{section}

\section{Method}
\label{sec:method}

\subsection{Template search engine}
\label{sec:search_engine}
As well established, LLMs still perform poorly at generating valid chemical representations such as SMILES or SMARTS, making them unreliable for direct use in chemical design tasks\cite{edwards-etal-2022-translation, walters2024silly, jang2024}. Therefore, we envisioned a template-based approach that leverages the analytical strengths of LLMs to convert reaction patterns into textual descriptions. This template search engine builds on the template library of AiZynthFinder \cite{genheden2020aizynthfinder}, a collection of approximately 40,000 unique SMARTS-encoded transformations generated with RDChiral \cite{Coley2019}.
Each reaction template was converted into a one-sentence textual description reflecting the chemically most relevant transformation features using \verb|Claude-3.7-Sonnet| and subsequently embedded into a vector space using various embedding models. To minimize linguistic bias within the embedding space, the prompt was iteratively refined and strictly constrained to enforce a consistent structure of the description across all templates, thereby ensuring that clustering reflected chemical semantics rather than syntactic similarity. 
% The resulting embeddings were benchmarked across multiple models (see Table XXX SI) using Top-\textit{k} retrieval based on cosine similarity as alternative distance metrics showed no significant performance difference. Evaluation was carried out on the retrieval rate of (i) expected semantic templates, (ii) chemically applicable templates and (iii) templates satisfying both criteria for different sizes of the template library. \verb|text-embeddings-ada-002| outperformed the other embedding models and consequently was used for the description of the entire AiZynthFinder library. 

\subsection{Phase 1: Initial planning by LLMs}

The workflow of Synthelite consists of two phases. The first phase aims to generate an initial synthesis route which serves as a blueprint for the second phase, in which MCTS is employed to locally sample actions that resemble those in the blueprint.

The blueprint route in the first phase is constructed by the LLMs in a greedy stepwise manner (see Section \ref{sec:strat_prompt}), iteratively from the target to the starting materials.
At each step, the LLMs are input with the target molecule, the user's prompt, and the current state of the synthesis, consisted of previous reactions and the leaf molecules (Figure \ref{prompt:input}). They are then asked to do the following tasks:
\begin{itemize}
    \item \textbf{Task 0: }Analyze the current state of the synthesis and decide whether to terminate (Figure \ref{prompt:task_0}. If a stop signal is captured, the following tasks would be skipped. It is noted that we give the LLMs all leaf molecules along with their availability in the stock, and the LLMs can choose to expand an in-stock molecule if necessary, in cases where users require a reaction performed in the early stage or those of starting material constraint.
    \item \textbf{Task 1: }Propose an overall multi-step synthesis plan to synthesize the target, taking into account the previous retro transformation, if any (Figure \ref{prompt:task_1}).
    \item \textbf{Task 2: }Based on the strategy of task 1, describe the next retro-transformation in a textual format (Figure \ref{prompt:task_2}). The LLMs also have to select a molecule to apply the reaction, as well as the reaction site to prevent site ambiguity of reaction templates. Atom mapping of leaf molecules is also provided as input, and the models can output the reaction site as a tuple of atom indices.
\end{itemize}

The textual description of the reaction is then used as a query to search for reaction templates (see Section \ref{sec:search_engine}). Relevant templates are applied to the selected leaf molecule, from which resultant reactions are filtered based on the LLM-specified reaction site.
The remaining reactions, up to 20, are then fed back to the LLMs, which select a single reaction that aligns with the Task 2's description (Figure \ref{prompt:select_action}). These steps are iterated until a stop signal is captured in Task 0, or the maximum number of steps is reached.

In our experiments, we limit the number of steps to be 25. To get a diverse set of solutions in the end, we also repeat Phase 1 three times, with subsequent iterations augmented with feedback from preceding ones.
The feedback is generated by another call to the LLMs, addressing the overall strategy and the chemical feasibility of each step (see Section \ref{sec:eval_prompt}).

We experiment with different frontier LLMs for the first phase, alternating between \verb|Claude-4.5-Sonnet|, \verb|Gemini-2.5-Pro|, and \verb|GPT-5|.

\subsection{Phase 2: Similarity-based MCTS}

Each attempt from Phase 1 results in a sequence of reference reactions $r^{\text{ref}}=\{r^{\text{ref}}_d\}_{i=d}^D$ and their corresponding textual description $q=\{q_d\}_{d=1}^D$, where $d$ represent the depth of the reaction in a synthesis tree. 
The logit $z$ for a retro-reaction $r(t)$, whose template is $t$, at depth $d$ is calculated as:

\[
z(r(t)|d) = \alpha \, \text{sim}(t, q_d) + (1-\alpha)\frac{N(t)}{N(t) + C}
\]

where $\text{sim}(t, q_d)$ is the textual embedding similarity returned by the template search engine, $N(t)$ is the library occurrence of the template $t$, and $C$ is a scaling constant. $\alpha$ is a parameter that controls the balance between strategy alignment and reaction popularity. 
Since the reaction-site of a reaction template is ambiguous, we only enumerate reactions that have the same reaction site with the reference reaction $r^{\text{ref}}_d$.

In our experiments, we set $\alpha=0.5$ and $C=100$. For each pair of $r^{\text{ref}}$ and $q$ resulted from one attempt from Phase 1, MCTS was run for 300 iterations.
It should be noted that, since the expansion policy only depends on the current depth of the search tree, we can cache the expansion actions based on the current depth and only call the search engine during the first iteration.
Final routes are ranked based on their alignment with the Phase 1 blueprint, giving greater weight to routes originating from later Phase 1 attempts.
% Intuition: Phase 1 the LLM proposes a blue-print, while Phase 2 mix-and-matches similar reactions but with different functional groups to arrive at in-stock materials

\section{Experiment}

\subsection{Strategic synthesis benchmark}
\label{sec:benchmark}

\begin{figure}[h]
    \centering
    \includegraphics[width=0.93\linewidth]{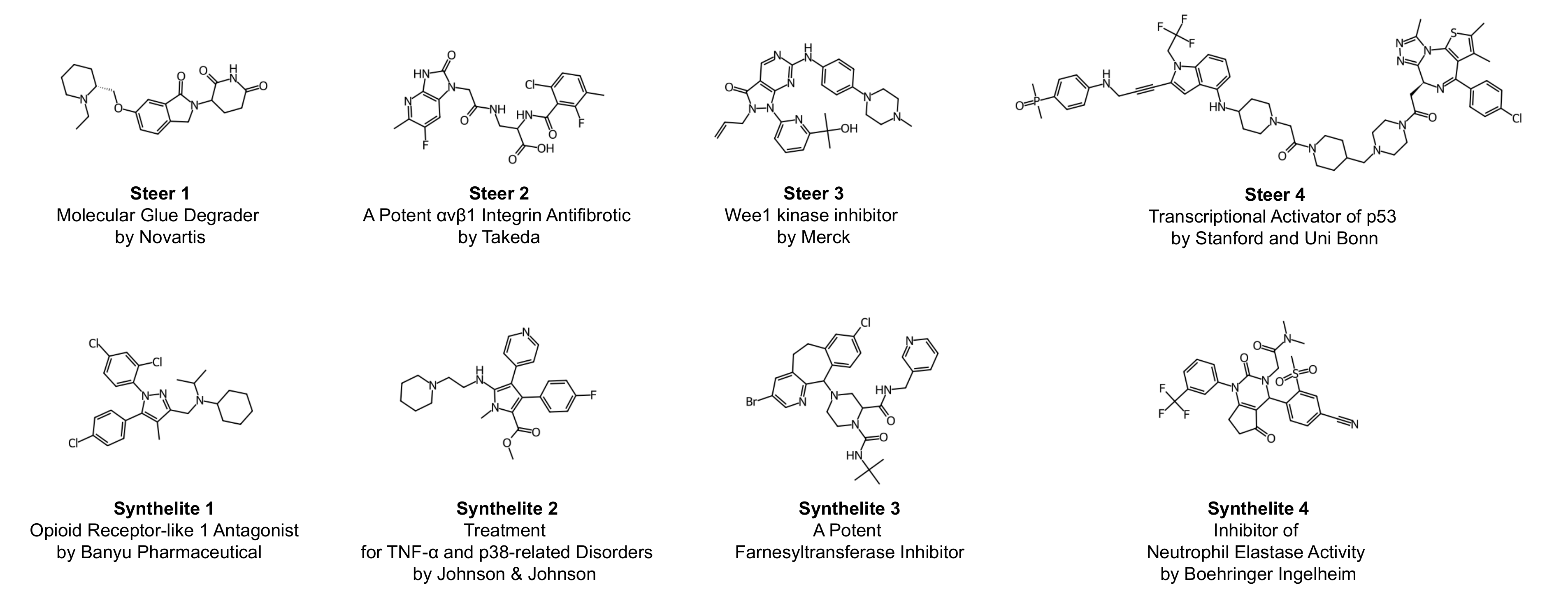}
    \caption{Target molecules used for benchmarking.}
    \label{fig:targets}
\end{figure}

The benchmark consists of pairs of target molecules and expert prompts. We employ a benchmark established by \citet{bran2025chemical}, containing four targets with different levels of complexity, denoted as \textbf{Steer 1-4}, and their corresponding prompts. 
Expanding on this benchmark, we add four additional targets (\textbf{Synthelite 1-4}). The structures of all molecules can be found in Figure \ref{fig:targets}.
Each molecular target is associated with a number of steering prompts written in textual format, ranging from reaction preference (such as performing a certain reaction at a specific stage in the synthesis) to a complex strategic sequence of reactions (perform reaction 1, then reaction 2, then reaction 3...). For a detailed list of specific prompts used for each target, see Tables \ref{tab:steer_prompt} and \ref{tab:Synthelite_prompt}.

For each pair of target-prompt in the benchmark, we manually curate a script to evaluate the route, using SMIRK pattern matching to identify the reaction pattern specified in the prompt. Given a retrosynthetic pathway written in \verb|json|, the functions would output a binary outcome, whether the route follows the prompt or not.
The evaluation scripts will be soon made available on github.

As the retrosynthesis software may provide multiple solutions for a case of target-prompt, we choose to quantify model performance using recall and precision (Figure \ref{fig:main_result}a). Recall, or case-wise success rate, measures the amount of cases where the model can find at least one route that passes the corresponding validation script.
Precision measures out of all the routes that the model suggests, how many of them satisfies the expert prompt. 
We envision that a good model would not just be able to find a solution for a case, but also return as many prompt-relevant solutions as possible.

For each target, we also run Synthelite with a neutral prompt to test the capability of the LLMs in synthesis planning without human biases:

\begin{minipage}{\linewidth}
\begin{tcolorbox}[
  colback=yellow!5!white,        % Background color
  colframe=yellow!75!black,      % Frame color
  title=Neutral prompt,    % Header/title of the box
  fonttitle=\bfseries,         % Bold title font
  sharp corners,               % Sharp corners
  boxrule=0.8pt                % Thickness of the frame
]
Highly feasible synthesis with high overall yields, consider potential side reactions and byproducts. Also ensure no unnecessary reactions are performed.
\end{tcolorbox}
\captionof{figure}{Neutral prompt that emphasizes on chemical feasibility.}
\label{prompt:neutral}
\end{minipage}

% \usepackage{booktabs}
% \usepackage{multirow}
% \usepackage{booktabs}
% \usepackage{multirow}
% \usepackage{graphicx}
\begin{table}[h!]
    \centering
    \caption{Prompts for Steer 1-4 targets.}
    \label{tab:steer_prompt}
    % \resizebox{\textwidth}{!}{%
    \begin{tabular}{@{} l c p{0.7\textwidth} @{}}
    \toprule
    Target ID & Prompt ID & Prompt \\ \midrule
    \multirow{8}{*}{Steer 1} & A & Break the piperidine and oxoisoindolinone rings in the retrosynthesis. Get the piperidine-2,6-dione ring from commercially available materials. \\
     & B & Break piperidine-2,6-dione and oxoisoindolinone rings in the retrosynthesis. Get the other piperidine ring from commercially available materials. \\
     & C & Break only oxoisoindolinone ring in retrosynthesis. Get piperidine-2,6-dione and piperidine rings from commercially available materials. \\ \midrule
    \multirow{3}{*}{Steer 2} & A & No ring formation reaction. \\
     & B & Late imidazole ring formation. \\
     & C & Early imidazole ring formation. \\ \midrule
    \multirow{3}{*}{Steer 3} & A & Do not break any ring but get all rings from commercial materials. \\
     & B & Break pyrimidine in the early stage but get all other rings from commercially available materials. \\ \midrule
    \multirow{16}{*}{Steer 4} & A & Identify the disconnection strategy that will cut the molecule in two similarly sized intermediates. The disconnection should be made between two piperidine rings. \\
     & B & Identify the disconnection strategy where the key disconnection will be made between indole and amino-piperidine rings. \\
     & C & Identify the disconnection strategy that will cut the molecule in two similarly sized intermediates. The disconnection should be made between piperazine and piperidine rings. \\
     & D & Identify the disconnection strategy that will cut the molecule in two similarly sized intermediates. One intermediate will have piperidine, indole, and aniline rings. The other intermediate will have thiophenol, chlorobenzene, diazepine, triazole, piperazine, and the other piperidine rings. \\
     & E & Identify the disconnection strategy that will cut the molecule in two intermediates. The disconnection should be made between diazepine and piperazine rings. \\ \bottomrule
    \end{tabular}%
    % }
\end{table}

\begin{table}[h]
    \centering
    \caption{Prompts for Synthelite 1-4 targets.}
    \label{tab:Synthelite_prompt}
    % \resizebox{\textwidth}{!}{%
    \begin{tabular}{@{} l c p{0.7\textwidth} @{}}
    \toprule
    Target ID & Prompt ID & Prompt \\ \midrule
    \multirow{4}{*}{Synthelite 1} & A & Late-stage pyrazole formation. \\
     & B & Early-stage pyrazole formation. \\
     & C & Pyrazole formation and late stage suzuki coupling. \\ 
     & D & Heterocycle formation, then amine coupling then Suzuki coupling. \\ \midrule
    \multirow{4}{*}{Synthelite 2} & A & Reductive amination in the final three reactions. \\
     & B & Early stage pyrrole formation. \\
     & C & Early imidazole ring formation. \\
     & D & Late stage fluorination. \\ \midrule
    \multirow{4}{*}{Synthelite 3} & A & Late stage amide coupling involving piperazine. \\
     & B & Late stage piperazine formation through intramolecular cyclisation. \\
     & C & Ensure a chlorobenzene fragment is preserved from the deepest starting material to the final product. \\ \midrule
    \multirow{5}{*}{Synthelite 4} & A & Late stage aldol condensation. \\
     & B & Late trifluromethylation. \\
     & C & Preserve methylsulfonyl group in all steps. \\
     & D & Final-state introduction of methyl sulfonyl group. \\
     & E & Heterocycle formation then SNAr reaction. \\ \bottomrule
    \end{tabular}%
    % }
\end{table}

\subsection{Starting-material constrained synthesis benchmark}
\label{sec:sm_constrained}

Due to the limitations of time and API budget, we tested the models on 20 randomly sampled pairs of target and building blocks from 140 targets of the Pistachio Reachable benchmark \cite{yu2024double}.
To guide the search toward the desired starting materials, we use the prompt template as shown in Figure \ref{prompt:sm_constraint}.
The synthesis planning attempt is considered successful if the constrained building block is found in the leaf molecules or intermediates.

\subsection{Baseline}

As a baseline, we compare Synthelite against the neural-guided synthesis planning component of AiZynthFinder \cite{genheden2020aizynthfinder}, which employs a single-step policy network trained on reaction data from the USPTO dataset \cite{Lowe2017}. Throughout this paper, we refer to this baseline simply as AiZynthFinder (AZF). As AZF does not support user-specified constraints or additional planning requirements, we evaluate it solely on the unconstrained benchmarking targets.

For the task of starting-material–constrained synthesis, we additionally include Tango* \cite{armstrong2025tango} as a baseline, as it is explicitly designed for this task. Tango* guides the search using the TANimoto Group Overlap (TANGO) score \cite{guo2024takes}, which quantifies the similarity between nodes in the synthesis tree and the enforced starting materials.

\begin{minipage}{\linewidth}
\begin{tcolorbox}[
  colback=yellow!5!white,        % Background color
  colframe=yellow!75!black,      % Frame color
  title=Starting material prompt,    % Header/title of the box
  fonttitle=\bfseries,         % Bold title font
  sharp corners,               % Sharp corners
  boxrule=0.8pt                % Thickness of the frame
]
The synthesis must originate from the starting material \{\{IUPAC\_NAME\}\}, which has the SMILES string \{\{SMILES\}\}. This specified starting material must be used in its entirety. It represents a terminal leaf node in the retrosynthetic tree and must not be disconnected further. Your proposed route should be strategically designed to incorporate this building block efficiently.
\end{tcolorbox}
\captionof{figure}{Prompt for starting materials constrained synthesis.}
\label{prompt:sm_constraint}
\end{minipage}

\section{System prompts}
This section describes the prompts given to the LLMs. Variables wrapped inside double pairs of curly brackets (\verb|{{| and \verb|}}|) are to be replaced with actual values.

\subsection{Reaction template description prompts}

This prompt was given to \verb|Claude-3.5-Sonnet| when we constructed the template search engine as discussed in Section \ref{sec:search_engine}.

\begin{minipage}{\linewidth}
\begin{tcolorbox}[
  colback=violet!5!white,        % Background color
  colframe=violet!75!black,      % Frame color
  title=Role description,    % Header/title of the box
  fonttitle=\bfseries,         % Bold title font
  sharp corners,               % Sharp corners
  boxrule=0.8pt                % Thickness of the frame
]
You are an expert chemist tasked with analyzing and describing a reaction template. Your goal is to interpret a provided SMARTS notation and create a concise, single-sentence description of the reaction.
\end{tcolorbox}
\captionof{figure}{Role and task description for the reaction template description.}
\label{prompt:role_template_describe}
\end{minipage}

\begin{minipage}{\linewidth}
\begin{tcolorbox}[
  colback=violet!5!white,        % Background color
  colframe=violet!75!black,      % Frame color
  title=Input specification,    % Header/title of the box
  fonttitle=\bfseries,         % Bold title font
  sharp corners,               % Sharp corners
  boxrule=0.8pt                % Thickness of the frame
]
Here is the reaction template in SMARTS format:\\
\\
\textless{}smarts\_reaction\textgreater{}\\
\{\{SMARTS\_REACTION\}\}\\
\textless{}/smarts\_reaction\textgreater{}\\
\\
which is written in the forward direction, with the reactant fragments are on the left side and the products are on the right side of the reaction arrow '\textgreater{}\textgreater{}'.
\end{tcolorbox}
\captionof{figure}{Input specification for template description.}
\label{prompt:input_template_describe}
\end{minipage}

\begin{minipage}[t]{\linewidth}
\begin{tcolorbox}[
  colback=RedViolet!5!white,        % Background color
  colframe=RedViolet!95!black,      % Frame color
  title=Instruction for template description,    % Header/title of the box
  fonttitle=\bfseries,         % Bold title font
  sharp corners,               % Sharp corners
  boxrule=0.8pt                % Thickness of the frame
]
Please conduct a thorough analysis of the reaction in \textless{}smarts\_breakdown\textgreater{} tags. Follow these steps:\\
...
\\
After completing your analysis, evaluate if this reaction template represents a chemically plausible transformation or not. If not, simply output:\\
\\
\textless{}description\textgreater{}\\
"This reaction template represents a chemically implausible transformation."\\
\textless{}/description\textgreater{} \\
\\
If it is a valid reaction, formulate a single-sentence description of the reaction in \textless{}description\textgreater{} tags. Your description should follow this structure:\\
\\
\textless{}description\textgreater{}\\
"This reaction involves [general transformation type in forward direction] ([specific name reaction(s)]) of [key reactant(s)] to form [key product(s)], focusing on the [key functional group(s)], and is classified as [reaction type(s)]."\\
\textless{}/description\textgreater{}\\
\\
Ensure that your description:\\
1. Is written in forward perspective (from reactants to products).\\
2. Includes all matching reaction types as possibilities if they differ only in mechanism or conditions but not in the outcome of the transformation.\\
3. Is coherent and clear about the direction of the reaction.\\
4. Specifies both the general transformation type and the specific name reaction(s). \\
5. If there are multiple name reactions which pass by fundamentally different mechanisms also consider those multiple reaction types in your description.
\end{tcolorbox}
\captionof{figure}{Instruction for template description.}
\label{prompt:instruction_template_describe}
\end{minipage}

\begin{minipage}{\linewidth}
\begin{tcolorbox}[
  colback=RedViolet!5!white,        % Background color
  colframe=RedViolet!95!black,      % Frame color
  title=Examples for template description,    % Header/title of the box
  fonttitle=\bfseries,         % Bold title font
  sharp corners,               % Sharp corners
  boxrule=0.8pt                % Thickness of the frame
]
Here are two examples of a description which you should not copy but use as a reference:\\
\\
C-O-C(=O)-[n;H0;D3;+0:1](:[\#7;a:2]):[c:3]\textgreater{}\textgreater{}[\#7;a:2]:[nH;D2;+0:1]:[c:3]\\
\textless{}description\textgreater{}\\
This reaction involves a protecting group removal (carbamate deprotection) of an N-methoxycarbonyl protected aromatic nitrogen heterocycle to form the corresponding NH-heterocycle, focusing on the N-CO2Me to N-H  transformation, and is classified as a nitrogen deprotection reaction.\\
\textless{}/description\textgreater{}\\
\\
Br-[CH2;D2;+0:1]-[C;H0;D3;+0:2]=O.[NH2;D1;+0:3]-[C;H0;D3;+0:4](=[S;H0;D1;+0:5])-[SH;D1;+0:6]\textgreater{}\textgreater{}\\
\text{[SH;D1;+0:6]-[c;H0;D3;+0:4]1:[n;H0;D2;+0:3]}\\
\text{:[c;H0;D3;+0:2]:[cH;D2;+0:1]
:[s;H0;D2;+0:5]:1}\\
\textless{}description\textgreater{}\\
This reaction involves a heterocycle formation (modified Hantzsch thiazole synthesis) of $\alpha$-bromo aldehyde and a thiourea derivative to form a thiol-substituted thiazole, focusing on the formation of the thiazole ring system through C-N and C-S bond formations, and is classified as a nucleophilic substitution-cyclization reaction.\\
\textless{}/description\textgreater{}
\end{tcolorbox}
\captionof{figure}{Examples for template description.}
\label{prompt:example_template_describe}
\end{minipage}

\subsection{Step-by-step synthesis planning prompts}
\label{sec:strat_prompt}

% \paragraph{Role and task description}

\begin{minipage}{\linewidth}
\begin{tcolorbox}[
  colback=violet!5!white,        % Background color
  colframe=violet!75!black,      % Frame color
  title=Role and task description,    % Header/title of the box
  fonttitle=\bfseries,         % Bold title font
  sharp corners,               % Sharp corners
  boxrule=0.8pt                % Thickness of the frame
]
You are an expert organic chemist assistant tasked with developing a retrosynthesis strategy for a target molecule. Your goal is to create a detailed synthesis plan that adheres to the user's prompt or constraints while following efficient and feasible retrosynthesis routes. From the synthesis plan that you produce, combined with the current state of the synthesis, you will also need to describe the next steps in the retrosynthesis. You also have access to search engine, by which you can search for reaction templates in a database using the description you produce. Since the search engine might return irrerelevant reactions, you also need to evaluate and select reactions that are chemically sound and aligned with your description.
\end{tcolorbox}
\captionof{figure}{Role and task description for the first phase of Synthelite.}
\label{prompt:role}
\end{minipage}

\begin{minipage}[t]{\linewidth}
\begin{tcolorbox}[
  colback=violet!5!white,        % Background color
  colframe=violet!75!black,      % Frame color
  title=Input specification,    % Header/title of the box
  fonttitle=\bfseries,         % Bold title font
  sharp corners,               % Sharp corners
  boxrule=0.8pt,                % Thickness of the frame
  listing only, 
  listing options={basicstyle=\ttfamily, breaklines=true}
]
Here is the target molecule for which you need to develop a synthesis strategy:\\
\\
\textless{}target\_molecule\textgreater{}\\
\{\{TARGET\_MOLECULE\}\}\\
\textless{}/target\_molecule\textgreater{}\\
\\
The user has provided the following prompt or constraint for this synthesis:\\
\\
\textless{}user\_prompt\textgreater{}\\
\{\{USER\_PROMPT\}\}\\
\textless{}/user\_prompt\textgreater{}\\
\\
Unless explicitly stated otherwise, assume that the user prompt describes a forward synthesis objective. When generating the retrosynthesis plan, interpret and reverse the constraints accordingly to guide the design of a backward synthesis route.
\end{tcolorbox}
\captionof{figure}{Passing the target molecule and expert prompt}
\label{prompt:input}
\end{minipage}

\begin{minipage}[t]{\linewidth}
\begin{tcolorbox}[
  colback=violet!5!white,        % Background color
  colframe=violet!75!black,      % Frame color
  title=Previous attempts,    % Header/title of the box
  fonttitle=\bfseries,         % Bold title font
  sharp corners,               % Sharp corners
  boxrule=0.8pt,                % Thickness of the frame
  listing only, 
  listing options={basicstyle=\ttfamily, breaklines=true}
]
The following routes are the results of your previous attempts, with the reactions written in retro direction. Each attempt may also contains feedbacks from your colleague, which could help you improve your next proposal:\\
\\
\textless{}previous\_attempts\textgreater{}\\
\{\{PREVIOUS\_ATTEMPTS\}\}\\
\textless{}/previous\_attempts\textgreater{}\\
\\
If a solution is found in your previous attempts, you can try to experiment with a different approach in this run so we can have diverse solutions in the end. Otherwise, the failed attempts can give you some hints on what to avoid in this run.
\end{tcolorbox}
\captionof{figure}{Giving the LLMs information of their previous attempts at solving the target.}
\label{prompt:attempts}
\end{minipage}

\begin{minipage}[t]{\linewidth}
\begin{tcolorbox}[
  colback=violet!5!white,        % Background color
  colframe=violet!75!black,      % Frame color
  title=Current synthesis state,    % Header/title of the box
  fonttitle=\bfseries,         % Bold title font
  sharp corners,               % Sharp corners
  boxrule=0.8pt,                % Thickness of the frame
  listing only, 
  listing options={basicstyle=\ttfamily, breaklines=true}
]
In the current run, the following steps have been done so far, given as a list of retro-reaction in SMILES format:\\
\\
\textless{}previous\_reactions\textgreater{}\\
\{\{PREVIOUS\_REACTIONS\}\}\\
\textless{}/previous\_reactions\textgreater{}\\
\\
Note that the reactions are written in retro direction, with products on the left and reactants on the right of the reaction arrow '\textgreater{}\textgreater{}'.\\
An empty list of previous reactions means that no steps have been done so far, and the synthesis starts from the target molecule.\\
\\
These steps give rise to the following 0-indexed list of molecules that could be retrosynthetically expanded:\\
\\
\textless{}current\_molecule\textgreater{}\\
\{\{CURRENT\_MOLECULE\_SMILES\}\}\\
\textless{}/current\_molecule\textgreater{}\\
\\
To aid your analysis, here are the SMILES strings with atom indices of the expandable molecules:\\
\\
\textless{}current\_molecule\_mapped\textgreater{}\\
\{\{CURRENT\_MOLECULE\_SMILES\_MAPPED\}\}\\
\textless{}/current\_molecule\_mapped\textgreater{}\\
\\
NOTE: If the availability of the molecules in \textless{}current\_molecule\textgreater{} and \textless{}current\_molecule\_mapped\textgreater{} are provided, prioritize expanding the molecules that are not in-stock, one at a time. If all molecules are in-stock but the user-query is not yet satisfied, continue to choose and expand a molecule.
\end{tcolorbox}
% \captionof{figure}{}
\label{prompt:input}
\end{minipage}

\begin{minipage}[t]{\linewidth}
\begin{tcolorbox}[
  colback=RedViolet!5!white,        % Background color
  colframe=RedViolet!95!black,      % Frame color
  title=Task 0: Stop or continue,    % Header/title of the box
  fonttitle=\bfseries,         % Bold title font
  sharp corners,               % Sharp corners
  boxrule=0.8pt,                % Thickness of the frame
  listing only, 
  listing options={basicstyle=\ttfamily, breaklines=true}
]
TASK 0: Analyze the \textless{}current\_molecule\textgreater{} in terms of synthetic accessibility. If all molecules are in-stock and the user-prompt is satisfied, you can give a stop signal and skip the remaining tasks by giving the following output. You could also give a stop signal if you think that it is better to use an explorative search to find a solution, given that the molecule is simple enough, or you think that if continuing would violate user's prompt.\\
\\
\textless{}stop\_signal\textgreater{}\\
TRUE\\
\textless{}/stop\_signal\textgreater{}
\end{tcolorbox}
\captionof{figure}{Instruction to terminate the retrosynthesis when suitable.}
\label{prompt:task_0}
\end{minipage}

\begin{minipage}[t]{\linewidth}
\begin{tcolorbox}[
  colback=RedViolet!5!white,        % Background color
  colframe=RedViolet!95!black,      % Frame color
  title=Task 1: Propose an overall plan,    % Header/title of the box
  fonttitle=\bfseries,         % Bold title font
  sharp corners,               % Sharp corners
  boxrule=0.8pt,                % Thickness of the frame
  listing only, 
  listing options={basicstyle=\ttfamily, breaklines=true}
]
TASK 1: Develop a synthesis strategy based on the target molecule and user prompt, following the steps outlined below.\\
\\
1. Analyze the structure of the target molecule and the expandable ones, identifying key functional groups and structural features.\\
2. Analyze the previous attempts and their feedbacks, if any.\\
3. Explain which reactions have been done so far.\\
   - For each reaction, describe which disconnection was made or which functional groups were changed, and explain how this contributes to the overall alignment with the user prompt.\\
4. Outline a preliminary retrosynthesis strategy continued from the previous steps, working backwards from the expandable molecules to simpler precursors. Ensure that you maintain a consistent retrosynthetic approach throughout your analysis, and that you align with the user prompt and the feedbacks from previous attempts.\\
5. Estimate the number of remaining steps required for the synthesis, taking into account the user's constraints.\\
6. Identify potential challenges or key considerations in the synthesis.\\
7. Evaluate potential protecting groups and their application points in the synthesis.\\
   - List out all potential protecting groups that might be necessary.\\
\\
Based on your analysis, develop a detailed, step-by-step retrosynthesis plan specifying retrosynthetic disconnections. For each step, specify the transformation type (e.g., reduction, oxidation, addition, elimination) and briefly describe the resulted fragments.\\
...\\
Present your final retrosynthesis plan in the following JSON-compatible format within \textless{}synthesis\_plan\textgreater{} tags:\\
\\
\textless{}synthesis\_plan\textgreater{}\\
\{\\
  "target\_smiles": "SMILES of the target molecule",\\
  "expandable\_molecules": "List of SMILES of the expandable molecules",\\
  "user\_constraint": "The provided user prompt or constraint",\\
  "previous\_steps": [\\
    \{\\
      "step\_number": 1,\\
      "step\_reaction": "SMILES of the retro-reaction at step 1",\\
      "step\_description": "Description of the retro transformation"\\
    \},\\
    \{\\
      "step\_number": 2,\\
      "step\_reaction": "SMILES of the retro-reaction at step 2",\\
      "step\_description": "Description of the retro transformation"\\
    \},\\
    ...\\
  ],\\
  "strategy\_overview": "...",\\
  "step\_estimate": "...",\\
  "next\_steps": [\\
    \{\\
      "step\_number": n, \# n is continuing from the last step number in previous\_steps\\
      "step\_description": "Description of the retro transformation"\\
    \},\\
    ...\\
  ],\\
  "additional\_notes": "Any relevant explanations on the synthesis strategy"\\
\}\\
\textless{}/synthesis\_plan\textgreater{}
\end{tcolorbox}
\captionof{figure}{Instruction for task 1 - Proposing an overal synthesis plan.}
\label{prompt:task_1}
\end{minipage}

\begin{minipage}[t]{\linewidth}
\begin{tcolorbox}[
  colback=RedViolet!5!white,        % Background color
  colframe=RedViolet!95!black,      % Frame color
  title=Task 2: Describing next transformation,    % Header/title of the box
  fonttitle=\bfseries,         % Bold title font
  sharp corners,               % Sharp corners
  boxrule=0.8pt,                % Thickness of the frame
  listing only, 
  listing options={basicstyle=\ttfamily, breaklines=true}
]
TASK 2: Propose the next disconnection step in the retrosynthetic route based on the strategy you provided. You need to describe in details the retro-reaction corresponding with the next step in the synthesis plan that you propose. In the example JSON above, it would be step number n that you need to describe. Present your final output in the following format:\\
\\
\textless{}next\_retro\_transformation\textgreater{}\\
The next step in the retrosynthetic route involves [general transformation type in retro direction] to yield [general reactant class 1] and [general reactant class 2] from [general product class], focusing on [key functional group or structural feature].\\
\textless{}/next\_retro\_transformation\textgreater{}\\
\\
From the proposed transformation, describe the corresponding reaction in the forward direction (i.e. inverse of \textless{}next\_retro\_transformation\textgreater{}, for example if in \textless{}next\_retro\_transformation\textgreater{} you describe a protection reaction, then the forward reaction would be a deprotection reaction), following the format:\\
\\
\textless{}next\_forward\_reaction\textgreater{}\\
The corresponding reaction involves [general transformation type in forward direction] ([specific name reaction(s)]) of [key reactant(s)] to form [key product(s)], focusing on the [key functional group(s)], and is classified as [reaction type(s)].\\
\textless{}/next\_forward\_reaction\textgreater{}\\
\\
Also give the index of the expandable molecule (0-indexed) on which the disconnection is made:\\
\\
\textless{}expandable\_molecule\_index\textgreater{}\\
A single integer, e.g. 0\\
\textless{}/expandable\_molecule\_index\textgreater{}\\
\\
And the a tuple of atom indices (based on the \textless{}current\_molecule\_mapped\textgreater{}) indicating the atoms that are involved in the reaction:\\
\\
\textless{}reaction\_atom\_indices\textgreater{}\\
A list of integers describing ONLY the cleaved or added bonds. Typical length is 1 or 2. For example, if a C-N bond is cleaved, and the C has atom index 3 and N has atom index 7, then the output should be [3, 7].\\
\textless{}/reaction\_atom\_indices\textgreater{}
\end{tcolorbox}
\captionof{figure}{Instruction for task 2 - Describing the next retro transformation.}
\label{prompt:task_2}
\end{minipage}

\begin{minipage}[t]{\linewidth}
\begin{tcolorbox}[
  colback=RedViolet!5!white,        % Background color
  colframe=RedViolet!95!black,      % Frame color
  title=Select reactions returned from search engine,    % Header/title of the box
  fonttitle=\bfseries,         % Bold title font
  sharp corners,               % Sharp corners
  boxrule=0.8pt,                % Thickness of the frame
  listing only, 
  listing options={basicstyle=\ttfamily, breaklines=true}
]
The search engine has returned the following reactions written in forward direction based on the queried description:\\
\\
\textless{}search\_result\textgreater{}\\
\{\{SEARCH\_RESULT\}\}\\
\textless{}/search\_result\textgreater{}\\
\\
Select \{\{MAX\_SELECTS\_REACTIONS\}\} that are chemically sound and aligned with your description in TASK 2 and strategy in TASK 1, in terms of not just reaction type but also the reactant fragments. A reaction is acceptable as long as its disconnection aligned with the description in TASK 2, even if it is not an exact match (e.g. if you described a bromine substitution,  a reaction having iodine substitution would be still acceptable). Then, give your final selection in the following format:\\
\\
\textless{}selected\_reaction\_indices\textgreater{}\\
List of integers, e.g. [0, 2, 3], ranked by the alignment with TASK 2 description and reaction feasibility, with the most preferred one being the first.\\
\textless{}/selected\_reaction\_indices\textgreater{}
\end{tcolorbox}
\captionof{figure}{Instruction to select reactions returned from the search engine.}
\label{prompt:select_action}
\end{minipage}

\subsection{Self-evaluation prompt}
\label{sec:eval_prompt}

A self-evaluation is performed after each attempt during Phase 1, and the resulting feedback is incorporated as additional input in subsequent attempts.

\begin{minipage}{\linewidth}
\begin{tcolorbox}[
  colback=violet!5!white,        % Background color
  colframe=violet!75!black,      % Frame color
  title=Role and task description,    % Header/title of the box
  fonttitle=\bfseries,         % Bold title font
  sharp corners,               % Sharp corners
  boxrule=0.8pt                % Thickness of the frame
]
You are an expert organic chemist, together with your colleague, tasked with developing a retrosynthesis strategy for a target molecule with some constraints provided by an user. Your colleage has proposed a synthesis plan, and your task is to evaluate the plan based on the chemical feasibility of the listed reactions, and whether this synthesis plan satisfies the constraints provided by the user.
\end{tcolorbox}
\captionof{figure}{Role and task description for self-evaluation.}
\label{prompt:eval_role}
\end{minipage}

\begin{minipage}{\linewidth}
\begin{tcolorbox}[
  colback=violet!5!white,        % Background color
  colframe=violet!75!black,      % Frame color
  title=Input specification,    % Header/title of the box
  fonttitle=\bfseries,         % Bold title font
  sharp corners,               % Sharp corners
  boxrule=0.8pt                % Thickness of the frame
]
Here is the target molecule for which you need to develop a synthesis strategy:\\
\\
\textless{}target\_molecule\textgreater{}\\
\{\{TARGET\_MOLECULE\}\}\\
\textless{}/target\_molecule\textgreater{}\\
\\
The user has provided the following prompt or constraint for this synthesis:\\
\\
\textless{}user\_prompt\textgreater{}\\
\{\{USER\_PROMPT\}\}\\
\textless{}/user\_prompt\textgreater{}\\
\\
Your colleague has proposed the following step-by-step synthesis plan. Each reaction is written in the retro direction, i.e., from product to reactants.\\
\\
\textless{}proposed\_synthesis\_plan\textgreater{}\\
\{\{PROPOSED\_SYNTHESIS\_PLAN\}\}\\
\textless{}/proposed\_synthesis\_plan\textgreater{}
\end{tcolorbox}
\captionof{figure}{Input for self-evaluation.}
\label{prompt:input_eval}
\end{minipage}

\begin{minipage}[t]{\linewidth}
\begin{tcolorbox}[
  colback=RedViolet!5!white,        % Background color
  colframe=RedViolet!95!black,      % Frame color
  title=Instruction for self-evaluation,    % Header/title of the box
  fonttitle=\bfseries,         % Bold title font
  sharp corners,               % Sharp corners
  boxrule=0.8pt,                % Thickness of the frame
  listing only, 
  listing options={basicstyle=\ttfamily, breaklines=true}
]
Analyze each reaction in the proposed sequence. For each reaction:\\
\\
1. Identify the key functional groups and structural changes involved.\\
2. Assess the feasibility of the proposed transformation.\\
3. Consider possible side reactions or alternative outcomes.\\
4. Evaluate how well the reaction aligns with the query's requirements.\\
5. Discuss any potential issues or improvements.\\
\\
After analyzing all reactions, assess the overall relevance of the proposed synthetic route to the query. Consider:\\
\\
1. How well does the overall sequence align with the query's goals?\\
2. Are there any major discrepancies or missing steps?\\
3. Are there any unnecessary or overly complex steps?\\
\\
Give your analysis in the following format:\\
\\
\textless{}synthesis\_analysis\textgreater{}\\
Your analysis here, highlighting problematic steps, and overall quality of the proposed synthesis in terms of feasibility and relevance to the user prompt.\\
\textless{}/synthesis\_analysis\textgreater{}\\
\\
Based on the analysis, provide your colleage's feedback on what to improve in their next proposal. Be specific about which steps need revision and why, focusing only on the steps that are problematic. If you think that the synthesis plan is high quality and meets the user's needs, acknowledge that as well and encourage your colleague to try other approaches that are strategically different so in the end we could have more than one solutions to choose from.\\
\\
\textless{}feedback\textgreater{}\\
\{\\
    "overall\_feedback": \textless{}A short overall feedback on the whole procedure\textgreater{},\\
    "problematic\_steps": [\\
        \{\\
            "step\_id": \textless{}Step number\textgreater{},\\
            "feedback": \textless{}A brief feedback on the step\textgreater{}\\
        \}\\
        // ... more problematic steps if needed\\
    ]\\
\}\\
\textless{}/feedback\textgreater{}\\
\\
Remember, the reactions shown are theoretical and have not been tested in a laboratory. They represent desired transformations but may not necessarily reflect what would actually occur in a flask. Your expertise is crucial in assessing the feasibility and relevance of these proposed reactions.
\end{tcolorbox}
\captionof{figure}{Instruction for self-evaluation}
\label{prompt:select_action}
\end{minipage}

%%%%%%%%%%%%%%%%%%%%%%%%%%%%%%%%%%%%%%%%%%%%%%%%%%%%%%%%%%%%

\end{document}